\documentclass{l4dc2024}
\usepackage{graphicx}
\usepackage{amsmath}
\usepackage{amsfonts}
\usepackage{amssymb}
\usepackage{bbm}
\usepackage{color}
\usepackage{subcaption}

\usepackage{float}
\usepackage{bm}
\usepackage{tikz}
\usetikzlibrary{bayesnet}
\usepackage{booktabs}
\usepackage{tikz}
\usepackage{array}
\usepackage{verbatim}
\usepackage{csquotes}
\usepackage[capitalize,noabbrev,nameinlink]{cleveref}
\crefformat{equation}{(#2#1#3)}
\Crefname{algocfline}{Algorithm}{Algorithms}
\Crefname{algocf}{line}{lines}
\Crefname{assumption}{Assumption}{Assumptions}
\crefrangeformat{assumption}{Assumptions~#3#1#4-#5#2#6}
\Crefname{hypothesis}{Hypothesis}{Hypotheses}
\crefrangeformat{hypothesis}{Hypotheses~#3#1#4-#5#2#6}

\usepackage[acronyms]{glossaries}
\glsdisablehyper
\makeglossaries

\newcommand{\ac}[1]{\gls*{#1}}
\newcommand{\acp}[1]{\glspl*{#1}}
\newcommand{\Ac}[1]{\Gls*{#1}}
\newcommand{\Acp}[1]{\Glspl*{#1}}

\newacronym{drl}{DRL}{deep reinforcement learning}

\newacronym{rl}{RL}{reinforcement learning}
\newacronym{tsda}{TSDA}{time symmetric data augmentation}
\newacronym{mdp}{MDP}{Markov decision process}

\newacronym{mc}{MC}{Markov chain}
\newacronym{drmc}{DRMC}{dynamically reversible Markov chain}
\newacronym{sac}{SAC}{soft actor-critic}
\newacronym{da}{DA}{data augmentation} %
\newacronym{darmdp}{DARMDP}{dynamically action reversible Markov decision process}

\newcolumntype{C}{>{\centering\arraybackslash}m{2em}}

\usepackage{xcolor}
\newcommand{\david}[1]{{\color{magenta}[David: #1]}}

\newcommand{\brett}[1]{{\color{orange}[Brett: #1]}}

\renewcommand{\david}[1]{} 
\renewcommand{\brett}[1]{} 

\newcommand{\termfont}[1]{{\tt #1}}

\newcommand{\cmark}{\textcolor{green}{\checkmark}}
\newcommand{\xmark}{%
  \begin{tikzpicture}[baseline=0.0ex, line width=1.2pt, scale=0.2]
    \draw[red] (0,0) -- (1,1);
    \draw[red] (0,1) -- (1,0);
  \end{tikzpicture}%
}

\newtheorem{assumption}{Assumption}
\newtheorem{hypothesis}{Hypothesis}

\title{An Investigation of Time Reversal Symmetry in Reinforcement Learning}

\coltauthor{%
 \Name{Brett Barkley} \Email{bbarkley@utexas.edu}
 \AND
 \Name{Amy Zhang} \Email{amy.zhang@austin.utexas.edu}
  \AND
 \Name{David Fridovich-Keil} \Email{dfk@utexas.edu}\\
 \addr The University of Texas at Austin, Austin, TX 78712
}

\begin{document}

\maketitle

\begin{abstract}

One of the fundamental challenges associated with \ac{rl} is that collecting sufficient data can be both time-consuming and expensive. In this paper, we formalize a concept of time reversal symmetry in a \ac{mdp}, which builds upon the established structure of \acp{drmc} and time-reversibility in classical physics. 
Specifically, we investigate the utility of this concept in reducing the sample complexity of reinforcement learning.
We observe that utilizing the structure of time reversal in an \ac{mdp} allows every environment transition experienced by an agent to be transformed into a feasible reverse-time transition, effectively doubling the number of experiences in the environment. To test the usefulness of this newly synthesized data, we develop a novel approach called \ac{tsda} and investigate its application in both proprioceptive and pixel-based state within the realm of off-policy, model-free \ac{rl}. Empirical evaluations showcase how these synthetic transitions can enhance the sample efficiency of \ac{rl} agents in time reversible scenarios without friction or contact. We also test this method in more realistic environments where these assumptions are not globally satisfied. We find that \ac{tsda} can significantly degrade sample efficiency and policy performance, but can also improve sample efficiency under the right conditions. Ultimately we conclude that time symmetry shows promise in enhancing the sample efficiency of reinforcement learning and provide guidance when the environment and reward structures are of an appropriate form for \ac{tsda} to be employed effectively.

\end{abstract}

\begin{keywords}%
  Reinforcement learning, sample-efficient learning, data augmentation, physics-informed machine learning%
\end{keywords}

\section{Introduction}
\glsresetall
    
In this paper, we investigate time reversal symmetry in a \ac{mdp} and apply it as \ac{tsda} in \ac{drl}. \Ac{drl} is a machine learning paradigm that uses deep learning techniques and an agent's interactions with its environment to learn a policy that maximizes rewards. However, the efficacy of \ac{drl}, particularly in the real world, often depends upon the ability to collect sufficient data, which can be challenging due to its high cost and/or limited availability. 

Time reversal symmetry is particularly promising to this end, because for every rollout of the policy a feasible reverse time trajectory can be generated. This trajectory provides counterfactual, and likely off-policy, experiences that normally would require additional interactions with the environment to obtain. This information can be utilized by downstream learning algorithms in a number of ways; in this paper, we investigate data augmentation since it is relatively algorithm-agnostic.
In particular, \ac{tsda} enables us to effectively double the amount of experiential data in an off-policy manner. 

The central hypothesis of this work is that utilizing time symmetric data can improve sample efficiency in reinforcement learning if the underlying environment is time reversible. 
To acquire this data, we first provide explicit conditions that allow an observed state transition to be reversed in time through a conjugate transformation. 
Although these requirements are strict, and are readily violated by physical phenomena such as friction and impulsive contact forces,
we find empirically that time symmetric transitions can still remain valuable even in environments that do not strictly satisfy these constraints. 
This evidence suggests that using \ac{tsda} will directly extend to a wide variety of real-world learning problems, in which physical environment dynamics are rarely precisely time symmetric. 

To evaluate the practical utility of the time reversal phenomenon, we conduct extensive experiments on various time symmetric and asymmetric \ac{rl} benchmarks. These investigations show that using time reversibility for data augmentation can improve sample efficiency in both ideal environments and in cases where optimal policies are incentivized to avoid time asymmetric transitions. However, in time asymmetric environments without this reward structure, we find that the synthetic transitions hinder learning due to their physical infeasibility. 
Overall, our investigations of this phenomenon and its application to data augmentation demonstrate that it can be useful when generating counterfactual transitions, combines the strengths of model-based and model-free techniques, and retains the simplicity and flexibility of model-free algorithms.
\section{Time Reversal, Symmetry, and Data Augmentation in Learning}
Recent related work has proposed using time reversal for machine learning in both high dimensional (pixel-based) and low dimensional (proprioceptive) state tasks, and the success of these methods motivates our work.
In particular, \cite{NairBFLK20} proposed a self-supervised approach to goal-conditioned \ac{rl} which exploits reverse-time trajectories.
Similarly, \cite{edwards2018forwardbackward} incorporate a learned backward-dynamics model which aids in the computation of these reverse-time trajectories. 
However, neither \cite{NairBFLK20} nor \cite{edwards2018forwardbackward} formally consider or exploit time symmetric state transitions.

Our work also builds upon that of \cite{grinsztajn2021there}, which defines a notion of degree-$N$ action reversibility, i.e., conditions under which an action can be ``undone'' with a sequence of $N$ further actions.
The \ac{darmdp} introduced in this paper can be seen as an adaptation of this concept. 
 Recent work has also proposed exploiting physical symmetry as a form of data augmentation \citep{Lin_2020, mavalankar2020goal}; however, these approaches often require an expert to exhaustively identify such symmetries. Many recent methods also focus on using symmetry for constrained learning, rather than data augmentation, to improve learning efficiency in \ac{drl}. These ideas tend to focus either on policy and loss symmetry constraints \citep{abdolhosseini2019learning, yu2018learning} or underlying neural network structure \citep{van2020mdp, yu2018learning}. Both approaches show great promise for situations where an optimal policy should ideally exhibit symmetry properties; however, these approaches still require substantial expert knowledge and the addition of specialized structures in the learning formulation.

 Finally, \ac{da} has been increasingly used in \ac{rl}, with examples including image transformations \citep{laskin2020reinforcement, yarats2020improving, kostrikov2021image, yarats2021mastering}, exploiting symmetry in proprioceptive state \citep{silvergo2016, Lin_2020}, and adding noise to proprioceptive states \citep{laskin2020reinforcement}. Data augmentation techniques that are independent of symmetry such as image transformations and state noise \citep{laskin2020reinforcement, kostrikov2021image} require minimal knowledge of the environment, in contrast symmetry exploiting methods, e.g., geometric or reflection symmetry \citep{Lin_2020, mavalankar2020goal}, which require a priori environment knowledge. 

\section{Background}
This section introduces four essential concepts for understanding time reversal symmetry in \acp{mdp}: the physical phenomenon of time reversal symmetry, reversible \acp{mc}, dynamically reversible \acp{mc}, and the structure of an \ac{rl} problem.
\subsection{Time Reversal Symmetry in Dynamical Systems}
\label{dyn_sys_trev}
In classical mechanics, the concept of time symmetry asserts that the evolution of a physical process remains equally feasible even when the direction of time is reversed. This concept is generally associated with conservative systems wherein total energy remains constant, e.g. for systems without frictional or impulsive contact forces. One way to describe this phenomenon is using the Hamiltonian of a dynamical system, $H$. The Hamiltonian takes as input position and momentum, $\mathbf{q},\mathbf{p} \in \mathbb{R}^n$, and can be used to generate the equations of motion that govern a system's evolution \citep{lamb1998time}:
\begin{equation}
\label{eqn:ham_diffeq}
\frac{\partial H}{\partial \mathbf{p}} = \frac{d\mathbf{q}}{dt}, \frac{\partial H}{\partial \mathbf{q}} = -\frac{d\mathbf{p}}{dt}.
\end{equation}
If the system modeled by $H$ is conservative, the Hamiltonian will remain constant as a function of time and be an even function with respect to the momentum vector: $H(\mathbf{q}, -\mathbf{p}) = H(\mathbf{q}, \mathbf{p})$.
Under these conditions, the physical relations in \cref{eqn:ham_diffeq} will remain unchanged under the time reversal transformation:
\begin{equation}
\label{eqn:dyntrev}
T: (\mathbf{q}, \mathbf{p}, t) \rightarrow (\mathbf{q}, -\mathbf{p}, -t).
\end{equation}
This invariance implies that any forward time trajectory for a conservative system can be reversed in time and momentum to generate an equally feasible trajectory. 
\subsection{Reversible \acp{mc}}
\label{subsec:revmc}
A stationary \ac{mc} is reversible if it satisfies the detailed balance condition \citep{kelly_reversibility_nodate}:
\begin{equation}
\label{eqn:detailed_balance}
P(s_t) P(s_{t+1}|s_t) = P(s_{t+1}) P(s_t|s_{t+1}), \,\, \forall\,\, s_t, s_{t+1},
\end{equation}
where $P(s_t)$ and $P(s_{t+1})$ are the equilibrium probabilities of states $s_t$ and $s_{t+1}$, respectively, and $P(s_{t+1}|s_t)$ and $P(s_t|s_{t+1})$ are the probability of transitions from states $s_t$ to $s_{t+1}$ and from $s_{t+1}$ to $s_t$. For a \ac{mc} to remain in equilibrium regardless of transition direction, these transition probabilities must be balanced for each pair of states.

\subsection{Time Reversal Symmetry in \acp{mc}}
\label{timerev_mc}
\Acp{drmc} are a discrete-time analog to time reversal symmetry in classical physics \citep{kelly_reversibility_nodate}. A stationary \ac{mc} can be transformed into a \ac{drmc} by incorporating the concept of conjugate states. Given a state $s_t$, we obtain its conjugate state using an involution $f$:
\begin{equation}
\label{eqn:conjinvol}
s_t^+ = f(s_t).
\end{equation}

For a \ac{mc} to be dynamically reversible, conjugate states must satisfy detailed balance:
\begin{equation}
\label{eqn:dynrev_detbal}
P(s_t) P(s_{t+1}|s_t) = P(s_{t+1}^+) P(s_t^+|s_{t+1}^+), \,\, \forall\,\, s_t, s_{t+1}.
\end{equation}
For physical systems, $f$ corresponds to the continuous-time transformation $T$ introduced in \cref{eqn:dyntrev}, which negates the subset of the state dimensions corresponding to momentum \citep{lamb1998time, bidir1996} and is generally known a priori. Applying $f$ to a vector of position and momentum results in $f(\mathbf{q},\mathbf{p}) = (\mathbf{q},-\mathbf{p})$, which is identical to the time reversal transformation in \cref{eqn:dyntrev}, but without the explicit time negation. 

\subsection{Reinforcement Learning}
\ac{rl} algorithms aim to learn a policy that maximizes the expected cumulative reward obtained through interactions with an environment \citep{Sutton1998}. The agent's interaction can be modeled as an \ac{mdp}, which is a tuple $(S, A, P, R, \gamma)$, where $S$ is the state space, $A$ is the action space, $P(s_t|s_{t-1}, a)$ is the probability of transitioning from state $s_{t-1}\in S$ to $s_t \in S$ given action $a \in A$, and $\gamma \in [0, 1]$ is a discount factor that determines the importance of future rewards. In the \ac{mdp}, each state transition yields a reward $r=R(s)$ and the goal of \ac{rl} is to find a policy $a \sim \pi(\cdot|s)$ that maximizes the expected cumulative reward: $\mathbb{E}_{s_0 \sim p_0, \pi}\left[\sum_{t=0}^{\infty} \gamma^t R(s_t, a_t)\right]$,
where $s_0$ is the initial state, $p_0$ is the initial state distribution, and $\pi$ is the policy.

\section{Dynamically Action Reversible \acp{mdp} and Time Symmetric Data}
\label{darmdp}
\glsreset{darmdp}
In this section, we present two of our contributions: the concept of a \ac{darmdp} and a corresponding approach for data augmentation in off-policy \ac{rl}.
\subsection{The Dynamically Action Reversible \ac{mdp}}
\label{spec_darmdp}

Consider a physical system such as a spacecraft following a fixed orbit, subject to a gravitational field. The system can be modeled with a dynamically reversible \ac{mc} due to its conservative nature, but when thrusters are used to change orbit, energy is no longer conserved due to the introduction of a non-conservative force. Nevertheless, the system can still be reversed in time. We can integrate this new notion of time reversibility into an \ac{mdp} using \acp{darmdp}.
\begin{definition}[\ac{darmdp}]
\label{def:act_detbal}
Let $A$ and $S$ be the action and state space of an \ac{mdp}, $a^+ \in A$ and $s^+\in S$ be conjugate transformations of the action $a$ and state $s$, respectively, and $f$ an involution on $A$. An \ac{mdp} is a \ac{darmdp} if 
$\forall ~ s_t, s_{t+1} \in S, \exists ~ f(a)=a^+$ s.t. $P(s_t|s_{t-1}, a) = P(s_{t-1}^+|s_t^+, a^+)$.

\end{definition}
To illustrate \cref{def:act_detbal}, consider OpenAI's Lunar Lander \citep{brockman2016openai} with $a\in A$ comprising control over the main booster and $s \in S$ representing the altitude of the craft. As the lander descends, the state-action transition is represented by the tuple $(s_{t-1}, a, s_t)$ and the time reversed trajectory is $(s_t^+, a^+, s_{t-1}^+)$. The first tuple represents the lander decelerating during descent under action $a$, whereas the second tuple represents the time reversed case where the lander instead accelerates during ascent under $a^+$. If the state space is constrained to cases where no contact is made between the lander and the moon, this \ac{mdp} is a \ac{darmdp}. However, if the state space includes situations where high impulse contact is made with the lunar surface there will be cases where the detailed balance condition in \cref{def:act_detbal} does not hold and the \ac{mdp} is not a \ac{darmdp}.

 As shown in the example of the lunar lander, for a \acp{darmdp}, a state-action transition is transformed elementwise into its conjugate counterpart using a state-action involution analogous to that introduced in \cref{eqn:conjinvol}: $(s_t^+,a^+, s_{t-1}^+) = f(s_{t-1}, a, s_t)$.

\subsection{Using Conjugate Transitions for Data Augmentation}
\label{meth_data_aug}

In this section we explore the use of time reversal symmetry in \acp{darmdp} as a form of data augmentation to aid in training an \ac{rl} agent. While investigating \acrfull{tsda} we make the following assumptions:
\begin{assumption}
\label{ass_one}
  The action is unchanged under time reversal, i.e. $a = a^+$. This ensures that the policy only needs to be queried once for a conjugate pair of transitions, but is otherwise only included for simplicity of exposition.
\end{assumption}
\begin{assumption}
\label{ass_two}
  We have a priori knowledge of the MDP reward function. Consequently, the conjugate transition's reward can be assessed directly. 
\end{assumption}

\ac{tsda} uses the time symmetry involution introduced in \Cref{spec_darmdp} to produce reverse time transitions for each policy rollout in the \ac{darmdp}. These reverse time rollouts are counterfactual, in the sense that they are generally off-policy due to differences between $\pi(a|s_{t-1})$ and $\pi(a|s_{t}^+)$. 
The pair of forward and reverse time states, actions, and rewards are then added to the agent's replay buffer and subsequently used to learn a policy via an off-the-shelf \ac{rl} algorithm.

The \ac{tsda} technique is especially useful because the conjugate state-action involution $f$ can be used to find conjugate transitions for both proprioceptive and pixel-based states with minor modification.
Consider first a proprioceptive state transition $(s_{t-1}, a_{t-1}, s_t)$ with reward $r_{t-1}$. A feasible conjugate transition can be generated by simply applying the state-action involution $f$ from \Cref{spec_darmdp}, with the action $a$ held constant per \cref{ass_two}, and reward given by $r^+_{t-1} = R(s^+_{t-1})$. In the case of pixel-based states, the transformation reverses the order of a sequence of pixel-based states without further modification, analogous to playing a video backward.

\section{Experimental Setup}

To evaluate the role of time symmetry in data augmentation for RL we conduct a series of numerical experiments. In particular, in \cref{results_analysis} we investigate the following hypotheses:
\begin{hypothesis}
\label{hype:1}
Utilizing time reversal symmetry for data augmentation can improve sample efficiency in reinforcement learning if the underlying environment is time reversible.
\end{hypothesis}
\begin{hypothesis}
\label{hype:2}
\ac{tsda} can lead agents to favor early exploitation over exploration in time symmetric environments, e.g., when important exploratory actions lie at the edge of the action space.

\end{hypothesis}
\begin{hypothesis}
\label{hype:3}
\ac{tsda} enhances sample efficiency in time asymmetric \acp{mdp} if the reward structure incentivizes policies to avoid time asymmetric transitions, e.g., when high-impulse contact is a feature of a poorly performing policy. 
\end{hypothesis}
\begin{hypothesis}
\label{hype:4}
\Ac{tsda} will not be useful when time reversal symmetry is frequently broken in substantial and unavoidable ways, e.g., when high-impulse contact is required for an optimal policy.\end{hypothesis}

\subsection{Environment Setup}
We use several off-the-shelf simulation environments from the DeepMind Control Suite \citep{tassa2018deepmind} and OpenAI Gym \citep{brockman2016openai} to benchmark \ac{tsda}. 
DeepMind's Cartpole swing-up is used for testing 
\cref{hype:1,hype:3} by varying the friction coefficient, whereas Lunar Lander and Walker walk/run are used to test \cref{hype:2,hype:4}, respectively. 
In order to increase the sample complexity of time symmetric benchmark tasks, we designed a customized environment based upon the work in
\cite{tassa2018deepmind}. This environment consists of a
fully actuated n-link manipulator where each joint actuator has a configurable maximum torque. 
To study the role of actuator authority in \cref{hype:2}, we consider three levels of torque limits in this environment: underpowered (\termfont{uM}), intermediate (\termfont{iM}), and overpowered (\termfont{oM}).
The task is to extend the manipulator arm fully upright,
with a reward structure identical to the \termfont{cart-k-pole} example introduced in \cite{tassa2018deepmind}.

\cref{tab:rl_environments} enumerates these environments and their time symmetric properties. 
We also include the episodic return value at which the task is considered to be solved. A larger discussion about rewards and solving tasks is included in \cref{app:rewards}.
\begin{table}[htbp]
\label{table:envs}
\centering
\resizebox{0.9\textwidth}{!}{%
\begin{tabular}{|l|c|c|c|}
\hline
\textbf{Environment} & \textbf{Friction} & \textbf{Contact} & \textbf{Episodic Return for Solution} \\ \hline
DeepMind Cartpole & \cmark & \xmark & $750$ \\ \hline
Lunar Lander  & \cmark & \cmark & $200$\\ \hline
DeepMind Walker Stand/Run & \cmark & \cmark & 800/600 \\ \hline
Custom N-link Manipulator & \xmark & \xmark & $800$ \\ \hline
\end{tabular}%
}
\vspace{0.05cm}
\caption{Summary of benchmark environments. Those with two checkmarks are time symmetric.}
\label{tab:rl_environments}
\vspace{-.75cm}
\end{table}

\subsection{Algorithmic Baselines and Evaluation Metrics}
In our empirical studies, we employ the \ac{sac} algorithm \citep{haarnoja2018soft}, with and without \ac{tsda}. When state is encoded as a sequence of images we use \ac{sac}+AE \citep{yarats2020improving}, which augments \ac{sac} with a regularized convolutional neural network (CNN) based encoder/decoder. For proprioceptive state, the autoencoder is removed and states are passed directly to the \ac{sac} algorithm as in \cite{haarnoja2018soft}.
 
During training, the policy is evaluated every $4000$ environment state transitions and the average return over $10$ episodes is computed. Training is repeated for $10$ random seeds with the mean and standard deviation for episodic return reported. Relative performance between \ac{sac} and \ac{tsda}+\ac{sac} is determined by comparing both the averaged episodic return at training convergence and the number of environment steps required to pass an episodic return solution threshold as shown in \cref{tab:rl_environments}.

\section{Results and Analysis}
\label{results_analysis}
\subsection{Improved Performance in Time Symmetric Environments}
\label{time_sym_envs}

\begin{figure}
    \centering
    \subfigure[]{\label{fig:pixel_cart_nofric}\includegraphics[width=.3\textwidth]{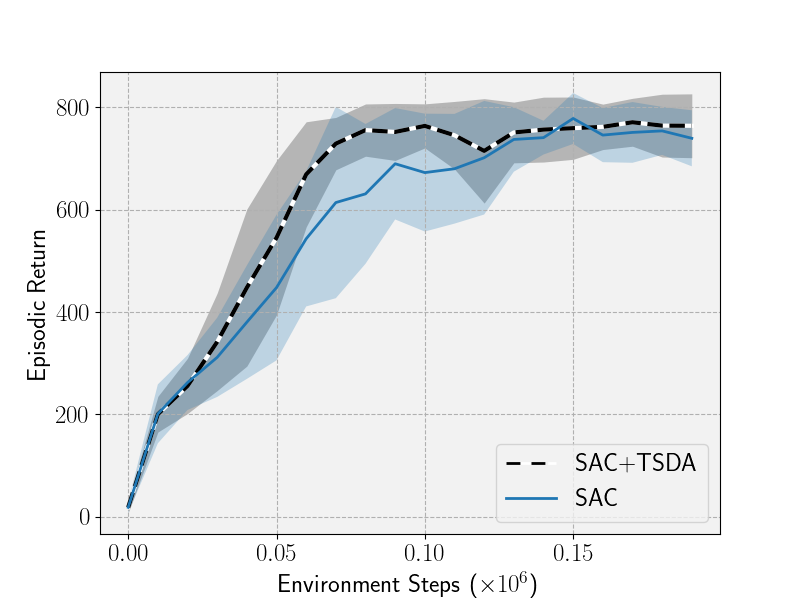}}
\hspace{1em}
        \subfigure[]{\label{fig:3linkonl}\includegraphics[width=.3\textwidth]{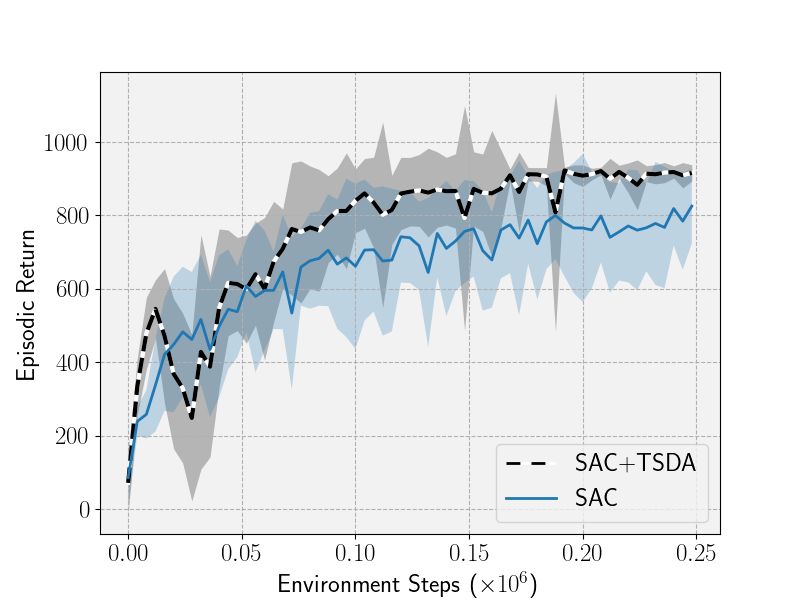}}
    \hspace{1em}
\subfigure[]{\label{fig:4linkiM}\includegraphics[width=.3\columnwidth]{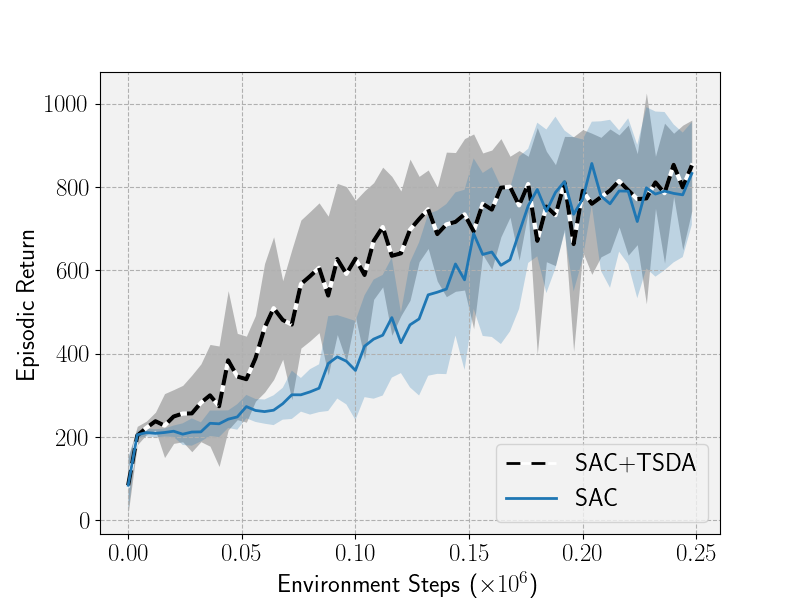}}
    \hspace{1em}

    \caption{\acp{mdp} where \ac{tsda} data augmentation definitively improves sample efficiency. Results are averaged over $10$ seeds with shading indicating one standard deviation away from the mean. a) Cartpole swing-up from pixels. b) 3-link \termfont{oM} environment. c) 4-link \termfont{iM}.}
    \label{fig:tsdaworking}
\end{figure}

We begin by showing three cases that support \cref{hype:1}. In these cases, \ac{tsda} is applied in a time symmetric environment and improves the sample efficiency of learning, as summarized in \cref{fig:tsdaworking}. 
\cref{fig:pixel_cart_nofric} shows learning progress for a frictionless cartpole swing-up using pixel-based state and the CNN encoder/decoder structure  of \cite{yarats2020improving}. 
By leveraging both reverse time-transitions in pixels and true environment steps, the agent is able to find a policy that solves the task with $50$\% fewer environment steps.

In \cref{fig:3linkonl}, we observe that in the early stages of training with \ac{tsda}, there is a noticeable increase in the variance of the evaluated policies for the 3-link overpowered manipulator environment (\termfont{oM}). 
However, as training progresses, \ac{tsda}+\ac{sac} effectively solves the task with $\sim$$62$\% fewer samples and reaches an averaged episodic return that is $\sim$$12$\% higher than \ac{sac}.
\ac{tsda} performs well in both cases because the environment is time symmetric and the agent has access to a large amount of actuator authority, which leads to increased diversity in the transitions augmented by \ac{tsda}. A lack of diverse transitions becomes particularly important for agents acting in environments with less actuator authority, which we shall study shortly. However, for \cref{fig:pixel_cart_nofric,fig:3linkonl}, the introduction of \ac{tsda} allows the agent to explore more of the environment without experiencing any additional physical transitions, and the agent subsequently exploits this additional information in policy learning. 

As mentioned in \cref{hype:2}, when actuator authority is reduced, important exploratory actions used in \termfont{oM} may become infeasible, which can lead to degraded performance in \ac{tsda}. 
This effect is illustrated in \cref{fig:4linkiM} for an \termfont{iM} with 4-links where the actuator authority reduction is accompanied by a reduction in the performance gap between \ac{sac} and \ac{tsda}+\ac{sac}. Specifically, the slope of the \ac{tsda}+\ac{sac} learning curve is significantly larger than \ac{sac} early in training, but both solve the task in the same number of samples. This effect implies that the time reversed transitions in this smaller action space are nearer to on-policy and less diverse, which incentivizes exploitation over exploration.
We provide evidence of this trend in \Cref{subopt_perf} where we consider a robotic system that has even further degradation in actuator authority.
\subsection{Time Symmetric Environments with Limited Actuator Authority}
\label{subopt_perf}

\begin{figure}
    \centering

        \subfigure[]{\label{fig:3linkum}\includegraphics[width=.4\textwidth]{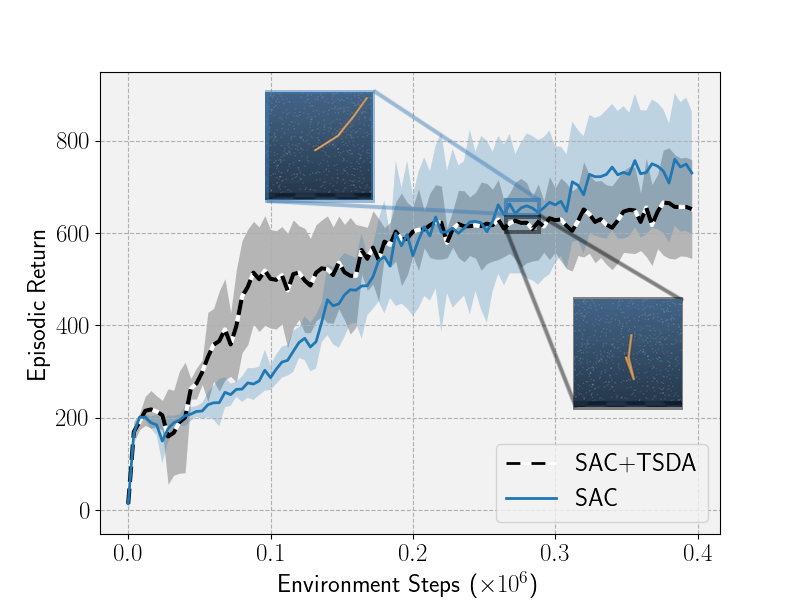}}
\hspace{1em}
    \subfigure[]{\label{fig:4linkum}\includegraphics[width=.4\textwidth]{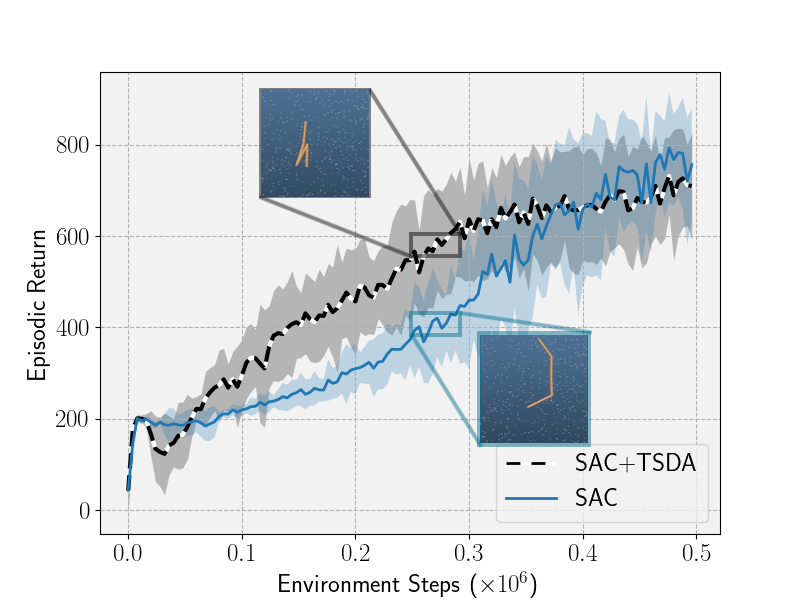}}
\hspace{1em}

    \caption{\ac{tsda} data augmentation leads to sub-optimal policies in training under limited actuator authority. a) 3-link \termfont{uM} environment. b) 4-link \termfont{uM} environment.}
    \label{fig:uMperf_mediocr}
\end{figure}

In this section we show two cases where an environment is time symmetric, yet \ac{tsda} does not improve sample efficiency in solving tasks.
\Cref{fig:3linkum,fig:4linkum} show cases of the n-link underpowered manipulator where the requirements of time symmetry are satisfied, but \ac{tsda} is harmful in achieving optimal policies. In both cases, \ac{sac}+\ac{tsda} achieves superior episodic returns before convergence, but \ac{sac} achieves higher final policy performance. 

These results suggest a deeper interaction between actuator authority and \ac{tsda}, as mentioned in \cref{hype:2}.
Here the exploration capabilities of the agent across a single environment transition are significantly degraded since it is operating in the \termfont{uM} environment. 
Consequently, the time-reversed transitions are even closer to on-policy than was the case for \cref{fig:4linkiM}, further biasing the agent toward exploitation.
This intuition is corroborated by both individual evaluation trajectories available during training and the recorded \href{https://github.com/CLeARoboticsLab/tsymRL}{video} of policy evaluation episodes. Snapshots of common states with and without \ac{tsda} are overlaid in \cref{fig:uMperf_mediocr} to illustrate this trend. In particular, we note that in the case of \ac{tsda}, these states correspond to sub-optimal outcomes that would likely require substantial exploration to escape. 
Ultimately, in both experiments we find that \ac{tsda} helps the agent identify these local optima faster and more frequently, but that their discovery hinders final policy performance.

\subsection{Improved Performance in Time \underline{A}symmetric Environments}
\label{subsec:workswhenshouldnt}
\begin{figure}
    \centering

            \subfigure[]{\label{fig:pixelcart}\includegraphics[width=.4\textwidth]{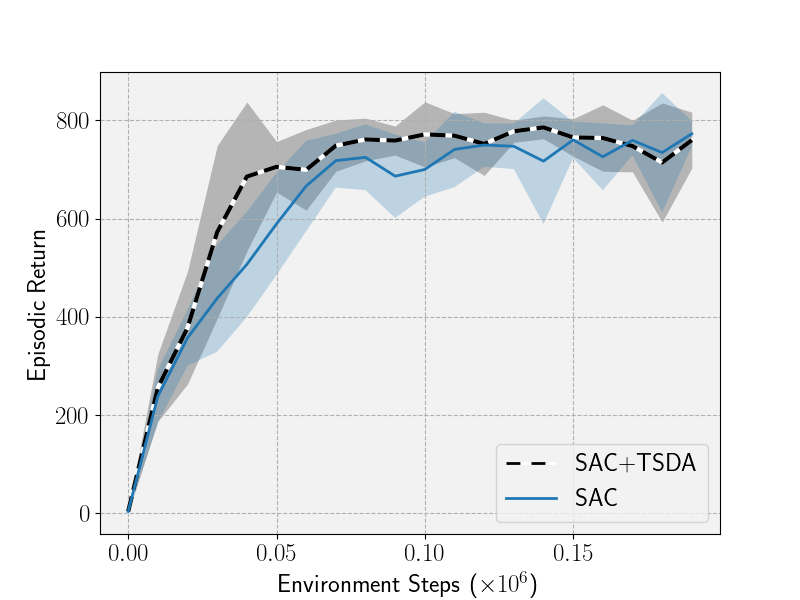}}
\hspace{1em}
        \subfigure[]{\label{fig:lunarlander}\includegraphics[width=.4\textwidth]{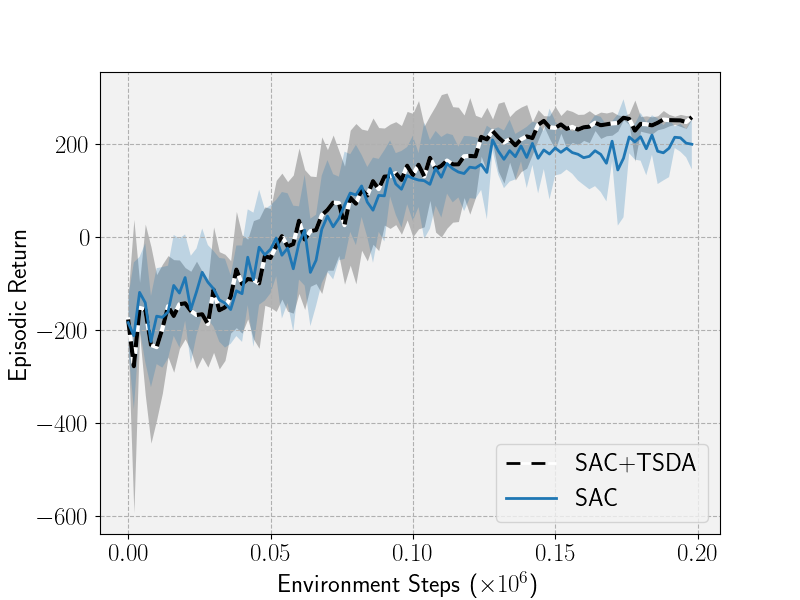}}
\hspace{1em}
    \caption{\ac{tsda} improves performance in some (not all) \underline{a}symmetric environments. a) Cartpole swing-up from pixels with high friction ($2000\times$ Mujoco defaults). b) OpenAI Gym Lunar Lander.}
    \label{fig:tsdashouldntbutdoes}
\end{figure}

\cref{fig:tsdashouldntbutdoes} shows two examples that support \cref{hype:3} wherein the uncontrolled \ac{mdp} is not strictly dynamically reversible, yet augmenting \ac{sac} with \ac{tsda} still improves performance. 
The first case, shown in \cref{fig:pixelcart}, is cartpole swing-up using pixel-based state; unlike the results of \cref{fig:pixel_cart_nofric} we now drastically increase the coefficient of friction to $2000\times$ its nominal value in \citep{tassa2018deepmind} and render the environment time asymmetric everywhere in the state space.
In such a case we would not expect \ac{tsda} to improve sample efficiency since transitions are universally time asymmetric, but instead we see that \ac{tsda}+\ac{sac}+AE still requires $\sim$$37$\% fewer samples to solve the task. 

One potential explanation for this phenomenon is that the CNN used to process image-based state in SAC+AE is unable to estimate the velocities in proprioceptive state accurately (cf. \citep{chua2018deep}), and it is these velocities which are most critical for encoding time reversal symmetry, per \cref{dyn_sys_trev}.
As a result, the frictional losses that make the system time asymmetric become difficult to detect when using images as state, making such environments good candidates for \ac{tsda}. 
\cref{fig:lunarlander} depicts policy training in the OpenAI Gym lunar lander environment \citep{brockman2016openai}. \ac{tsda}+\ac{sac} exhibits a higher sample efficiency relative to \ac{sac}, requiring approximately $27$\% fewer samples to solve the task and achieving a higher averaged episodic return. Initially, this outcome is surprising since the landing task requires making inelastic contact with the ground in the landing zone, which breaks time symmetry. However, this environment has a reward structure that incentivizes the policy to prioritize transitions that are time reversible. Specifically, the lander is considered to have crashed if its body makes contact with the ground and this results in a very negative reward. Consequently, optimal policies must ensure that the landing struts touch the ground first. Furthermore, the landing struts' attachment to the lander body is relatively fragile if subjected to excessive velocity/impulsivity. This inherent fragility creates an environment conducive to the benefits of \ac{tsda} since the optimal policy prioritizes minimal energy loss upon ground contact and thus avoids policies that will lead to violations of time symmetry in the \ac{mdp}.

\subsection{Poor Performance in Time Asymmetric Environments}
\begin{figure}
    \centering

                \subfigure[]{\label{fig:stand}\includegraphics[width=.4\textwidth]{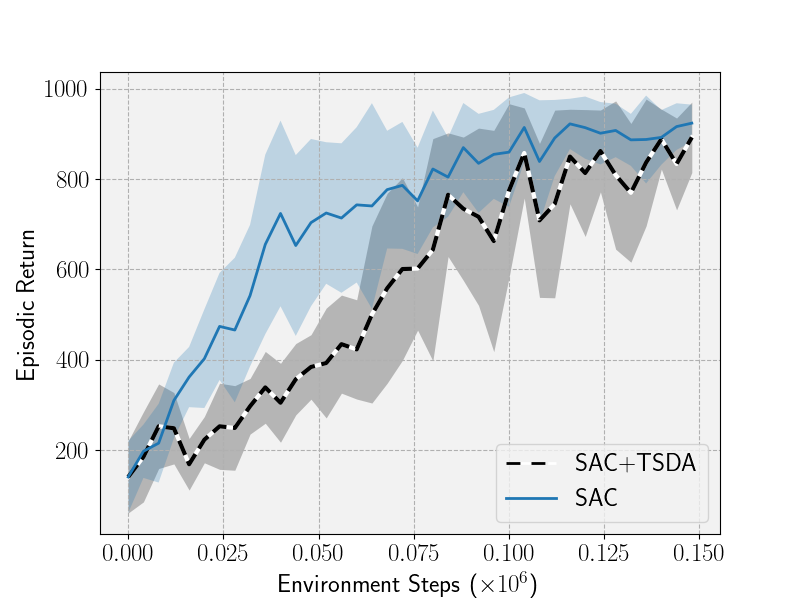}}
\hspace{1em}
            \subfigure[]{\label{fig:run}\includegraphics[width=.4\textwidth]{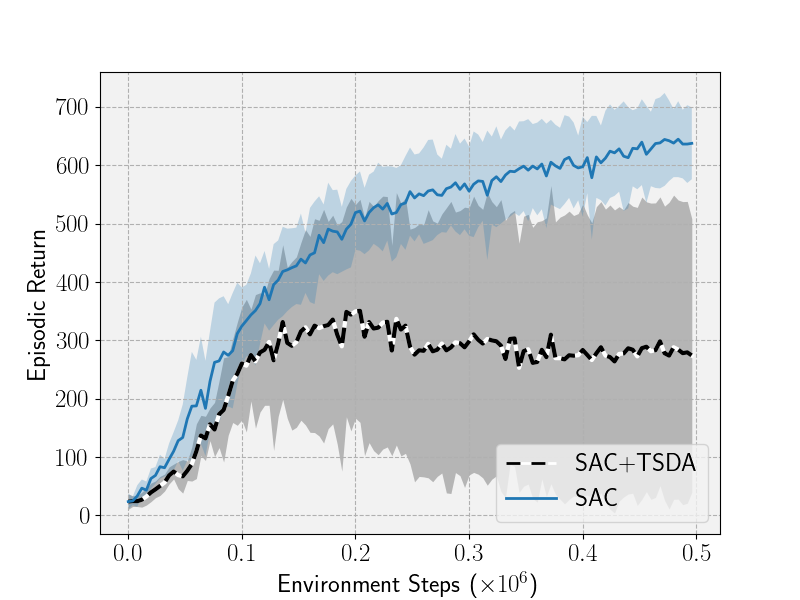}}
\hspace{1em}
    \caption{\ac{tsda} definitively worsens performance in (a) Walker \termfont{stand} (b) Walker \termfont{run}.}
    \label{fig:walker}
    \vspace{-.5cm}
\end{figure}

Finally, we investigate \cref{hype:4}, focusing on cases where \ac{tsda} leads to reduced sample efficiency and final policy performance due to time asymmetry in the environment. We then provide physical intuition to distinguish these cases from those of \cref{subsec:workswhenshouldnt} in which \ac{tsda} still improves policy performance. \cref{fig:walker} shows a comparison of policy training in the  Walker \termfont{stand} and \termfont{run} tasks from \cite{tassa2018deepmind}. In \cref{fig:stand}, \ac{sac}+\ac{tsda} requires $2\times$ more environment interactions to achieve similar levels of performance in averaged episodic return, but eventually solves the standing task. However, \ac{sac}+\ac{tsda} does not solve the \termfont{run} task despite early progress. 
Both of these tasks necessitate making high-impulse contact to attain high reward. In \termfont{stand}, falling is a frequent occurrence and invalidates the time reversed transitions from \ac{tsda} due to high impulse collisions. However, as an agent becomes capable fewer falls will occur and transitions will obey time reversal symmetry more frequently, explaining the eventual convergence of \ac{sac}+\ac{tsda} in \cref{fig:stand}. Similarly, the \termfont{run} task in \cref{fig:run} augments \termfont{stand}'s reward structure to encourage forward velocity. Policy performance in \termfont{run} requires competency in \termfont{stand}, but also frequent and high impulse collisions between the robot legs and the ground. As a result, the reward structure incentivizes operating in the time asymmetric portions of the \ac{mdp}, \ac{tsda} introduces many time reversed transitions that are infeasible, and policy training becomes erratic since many of the transitions are infeasible.
\label{tsdabad}

\section{Limitations and Conclusions}
In this work we formalize a notion of time reversal symmetry for \acp{mdp}, and investigate when and how it can be exploited to improve the sample efficiency of reinforcement learning.
We show that, when used for data augmentation in time symmetric environments, training reliably converges more quickly and can find better policies.
However, many real-world robotic environments are not time symmetric due to non-conservative forces, such as contact and friction, and the proposed method for \ac{tsda} can diminish performance in such cases (as illustrated in \cref{tsdabad}). 
Interestingly, however, we identify certain special cases which break the conditions of time symmetry, and in which \ac{tsda} still improves performance (cf. \cref{subsec:workswhenshouldnt}).
These examples provide evidence for the utility of this phenomenon in real-world robotics; however, an important limitation of this work is that we have yet to establish either a fundamental theoretical or real-world experimental understanding of these special cases. 
Future work should focus on establishing both theoretical and experimental results to this end.
Finally, generating time reversed transitions requires knowledge of the reward function during training; an assumption which is not strictly required in many \ac{rl} algorithms
\citep{Sutton1998}.


\bibliography{refs}
\section*{Appendix}
\addcontentsline{toc}{section}{Appendix} 

\appendix
\section{Hyperparameters and Setup}
 Although it no longer exhibits state of the art policy performance when using raw pixels \cite{kostrikov2021image, yarats2021mastering}, the SAC+AE \cite{yarats2020improving} implementation of SAC was chosen because it is easily configurable to either proprioceptive or pixel-based state. 
\subsection{Replay Buffer Size}
 The replay buffer employed alongside \acrshort{sac} is configured such that its capacity can hold the total number of training interactions the agent has with the environment. This is noteworthy because when incorporating \acrshort{tsda} with \acrshort{sac} it becomes necessary to include time-reversed transitions in the replay buffer. This addition effectively doubles the number of transitions per environment step and so we double the replay buffer capacity when using \ac{tsda} with \ac{sac}.
\subsection{Pixel Preprocessing}
    For simplicity and fairness of comparison, hyperparameters are held constant across tasks except in the case of the Walker environment and when using raw pixel states. In both these cases performance is very sensitive to action repeat \cite{yarats2020improving} and we used an action repeat of $2$. Image states are represented by three consecutive $84 \times 84$  RGB images stacked so inference about velocity and acceleration is possible as in \cite{mnih2013playing}.
    
\subsection{Other Hyperparameters}

\begin{tabular}{|l|l|}
\hline
\textbf{Parameter Name} & \textbf{Value} \\
\hline
Batch size & 128 \\
Discount $\gamma$ & 0.99 \\
Optimizer & Adam \\
Critic learning rate & $10^{-3}$ \\
Critic target updates per environment step & 2 \\
Critic Q-function soft-update rate & 0.01 \\
Critic encoder soft-update rate & 0.05 \\
Actor learning rate & $10^{-3}$ \\
Actor updates per environment step & 2 \\
Actor log stddev bounds & $[-10, 2]$ \\
Autoencoder learning rate & $10^{-3}$ \\
Temperature learning rate & $10^{-4}$ \\
Temperature Adam's $\beta_1$ & 0.5 \\
Initial temperature & 0.1 \\
\hline
\end{tabular}

\section{Time Reversible Integration in Discrete Time}
It is worth noting that the discussion of time reversal in \cref{dyn_sys_trev} is in continuous time, but \acp{mdp} are formulated in discrete time for \ac{rl}. This makes a short discussion of time reversibility in numerical integration important. To simulate the discretized dynamics in a time-symmetric way requires a symplectic integrator such as the Verlet integrator \citep{verlet_1967} and typical integrators used in many \ac{rl} benchmarks (e.g. fourth-order Runga-Kutta or Euler's method) do not have this property. In actuality the error induced by non-symplectic integrators is minimal unless integration is performed over long time horizons and we found it was safe to ignore without issue in our experimental evaluations.
\section{Task Rewards and Solutions}
\label{app:rewards}
\cref{tab:rl_environments} lists various environments along with their time symmetric properties and the episodic returns required to consider a task as solved. For tasks in the DeepMind Control Suite \citep{tassa2018deepmind}, the maximum episodic return is set at $1000$. However, achieving this maximum is uncommon in practice, leading to reported solutions with episodic returns significantly lower than $1000$ in prior research. These historical return values for task solutions within the DeepMind Control Suite are documented in \cref{tab:rl_environments}. In contrast, the OpenAI Gym \citep{brockman2016openai} explicitly defines a specific episodic return threshold as the criterion for successfully solving the Lunar Lander task \citep{lunarlander}. Finally, it should be highlighted that the N-link Manipulator environments were specifically developed for the purpose of testing our hypotheses, and thus, lack pre-existing literature regarding their solution criteria. Consequently, the episodic return threshold identified as a solution in \cref{tab:rl_environments} is based on the return value that yielded satisfactory outcomes in these tasks.

\end{document}